# ASSESSING THE IMPACT OF CNN AUTO ENCODER-BASED IMAGE DENOISING ON IMAGE CLASSIFICATION TASKS

Mohsen Hami[1,*], Mahdi JameBozorg[2]

*1 \** Bachelor's degree in Computer Engineering from Bu-Ali Sina University, Iran*

*2 undergraduate student* in Computer Engineering *at Bu-Ali Sina University, Iran*

## Abstract.

Images captured from the real world are often affected by different types of noise, which can significantly impact the performance of Computer Vision systems and the quality of visual data. This study presents a novel approach for defect detection in casting product noisy images, specifically focusing on submersible pump impellers. The methodology involves utilizing deep learning models such as VGG16, InceptionV3, and other models in both the spatial and frequency domains to identify noise types and defect status. The research process begins with preprocessing images, followed by applying denoising techniques tailored to specific noise categories. The goal is to enhance the accuracy and robustness of defect detection by integrating noise detection and denoising into the classification pipeline. The study achieved remarkable results using VGG16 for noise type classification in the frequency domain, achieving an accuracy of over 99%. Removal of salt and pepper noise resulted in an average SSIM of 87.9, while Gaussian noise removal had an average SSIM of 64.0, and periodic noise removal yielded an average SSIM of 81.6. This comprehensive approach showcases the effectiveness of the deep AutoEncoder model and median filter, for denoising strategies in real-world industrial applications. Finally, our study reports significant improvements in binary classification accuracy for defect detection compared to previous methods. For the VGG16 classifier, accuracy increased from 94.6% to 97.0%, demonstrating the effectiveness of the proposed noise detection and denoising approach. Similarly, for the InceptionV3 classifier, accuracy improved from 84.7% to 90.0%, further validating the benefits of integrating noise analysis into the classification pipeline.

## 1. Introduction

In the realm of Computer Vision, the accurate detection of defects in industrial products is crucial for ensuring quality control and efficiency in manufacturing processes. However, real-world images are often plagued by various types of noise, which can hinder the performance of defect detection algorithms. In this study, we focus on detecting defects in casting products, specifically submersible pump impellers, by leveraging deep learning models and innovative noise detection and denoising techniques. By addressing noise at its core and integrating it into the defect detection pipeline, with an end-to-end pipeline we aim to enhance the accuracy and reliability of defect identification in industrial settings. This research presents a comprehensive approach that not only identifies noise types but also effectively removes them to improve the accuracy of defect detection algorithms.

We have used the open source "casting product image data for quality inspection" dataset from Kaggle and received a noisy version of the dataset from the Bu-Ali Sina Department of Computer Engineering.

Our dataset comprises two phases. In the first phase, we have two classes (Defected and OK), with all images being noisy. The second phase includes noisy images along with their corresponding ground truth images. Our objective is to evaluate the performance of various deep learning models in binary classification tasks using noisy datasets (Phase 1). Additionally, we will devise an optimal approach for classifying noise types in the frequency domain and reducing noise using state-of-the-art Autoencoder models for Gaussian and periodic noise, as well as employing median filters with varying kernel sizes for salt and pepper noise types. These operations will be applied in Phase 2.

*Figure 1: comparing total params and sizes of different models by considering pre-trained weights and without it*

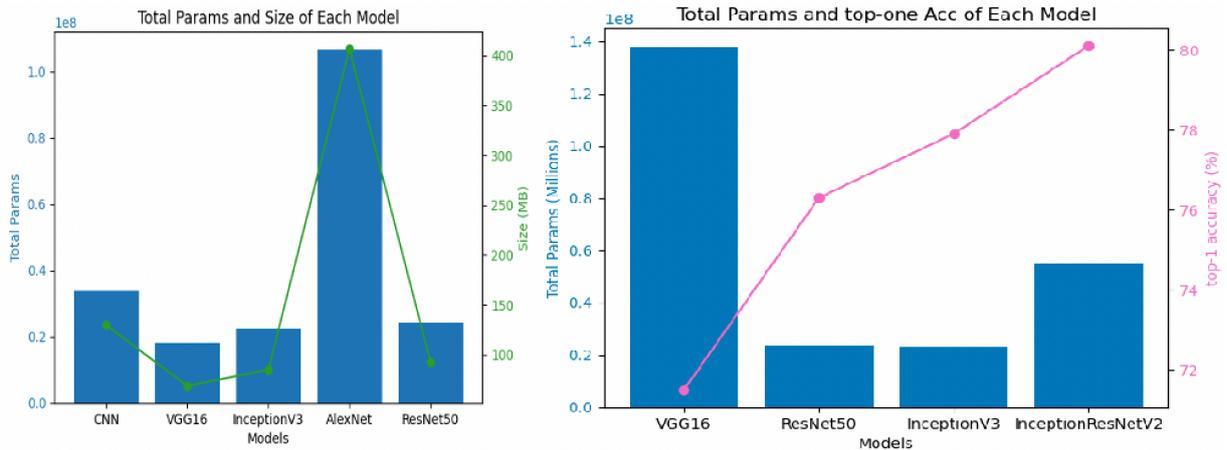

*Source: measurement of total params in Tensorflow and [1],[2],[3][4] and [5]*

Based on Figure 1 it's obvious the lightest model with pre-trained weights is VGG16 and the lightest model without pre-trained is our custom CNN architecture.

For tracing our work you can access both Phase1 and Phase2 datasets from these two links:

Phase 1, Phase 2, Labels

These links are taken from the computer engineering department of Bu-Ali Sina University.

## 2. Method

### 2.1 Binary classification

We have used different deep neural network architectures with transfer learning techniques to reach the best accuracy and generalization on the dataset. As you see, in the first phase we have two Defected and OK classes for casting products Which are noisy.

### 2.1.1 preprocessing

We divided our dataset into three separate parts, named Train, Test, and Validation sets, comprising approximately 70%, 15%, and 15% of the entire dataset, respectively. Subsequently, we randomly selected images to populate each set, resulting in 1063 images for the Train set, 85 images for the Test set, and 85 images for the Validation set. Additionally, because the input dimensions of our chosen neural network are 224 x 224, we resized all images accordingly and normalized them.

### 2.1.2 model selection and training

After completing the necessary preprocessing steps, the next step is to select a suitable model. In our study, we applied five deep neural networks (DNN) models under the same conditions to ensure comparability: VGG16, InceptionV3, AlexNet, ResNet50, and CNN, some of which come with pre-trained weights. To mitigate the risks of model overfitting and underfitting, we employed various callbacks in Keras, such as EarlyStopping. Additionally, we analyzed the trend of model accuracy using TensorBoard. [6]

2.1.3 model evaluation

We need to test the model's performance on unseen test instances, the results of the models are as shown below:

*Table 1: results of different models' binary classification metrics*

| Model | Train Accuracy | Test Accuracy | Test Loss | Precision | Recall | F1-score | Sensitivity | Specificity | #rank |
|---|---|---|---|---|---|---|---|---|---|
| CNN | 0.83 | 79.56 | 0.42 | 0.45 | 0.47 | 0.46 | 0.24 | 0.62 | 4 |
| VGG16 | 0.94 | 94.62 | 0.09 | 0.50 | 0.49 | 0.50 | 0.37 | 0.57 | 1 |
| ResNet50 | 0.64 | 68.23 | 0.61 | 0.54 | 0.56 | 0.54 | 0.27 | 0.75 | 5 |
| AlexNet | 0.82 | 81.17 | 0.40 | 0.47 | 0.44 | 0.44 | 0.45 | 0.42 | 3 |
| InceptionV3 | 0.91 | 84.70 | 0.37 | 0.56 | 0.54 | 0.55 | 0.54 | 0.53 | 2 |

*Source: outputs of classifier models*

Based on the final rank, we will choose two top models (VGG16 and InceptionV3) for our next study.

### 2.2 noise type detection

We classified different types of noises to enhance the model's performance and tailor our solution. For this purpose, we employed the VGG16 and Inception-ResNet-V2 classifiers in the frequency domain instead of the spatial domain. One of the primary reasons for utilizing the frequency domain in image classification tasks is its capability to distinguish between signals (desired information) and noises (unwanted artifacts), thereby enabling efficient image processing operations [7].

Fourier Transform is used to analyze the frequency characteristics of various filters. For images, 2D Discrete Fourier Transform (DFT) is used to find the frequency domain. A fast algorithm called Fast Fourier Transform (FFT) is used for the calculation of DFT.[8]

For a sinusoidal signal, $x(t)=A\sin(2\pi f t)$, we can say $f$ is the frequency of the signal, and if its frequency domain is taken, we can see a spike at $f$. If the signal is sampled to form a discrete signal, we get the same frequency domain but is periodic in the range $[-\pi,\pi]$ or $[0,2\pi]$ (or $[0,N]$ for N-point DFT). we can consider an image as a signal that is sampled in two directions. So taking Fourier transform in both X and Y directions gives us the frequency representation of the image.[8]

### 2.2.1 Preprocessing

We divided our dataset, as previously done, into Train, Test, and Validation sets, with proportions of approximately 60%, 20%, and 20%, respectively, relative to the entire dataset. Subsequently, we randomly selected images for each set, resulting in 750 images for training, 250 for testing, and 250 for validation. Additionally, since the input dimensions of our selected neural network are 224 x 224, we resized all images accordingly and converted them into the frequency domain.

### 2.2.2 Model selection and training

We utilized a pre-trained VGG16 and Inception-Resnet-V2, a state-of-the-art model for classification tasks, which significantly reduces the time required for model training and optimization [9]. We opted not to explore alternative architectures such as VGG19 or ResNet50, as research indicates their performance is inferior to VGG16 [10]

The Inception-ResNet-v2 model is chosen for its superior performance, efficiency, and versatility in computer vision tasks. Its combination of Inception and ResNet architectures allows for deep representation learning without the vanishing gradient problem, leading to state-of-the-art results in various applications while remaining computationally efficient.[5]

### 2.2.3 model evaluation

Finally testing these two models on unseen data we reached 99.5% after 24 epochs for VGG16 and 96.4% after 12 epochs for Inception-ResNet-v2.

So we have chosen VGG16 results for our future study.

*Figure 2: picture of a casting product image with Gaussian noise in the frequency domain*

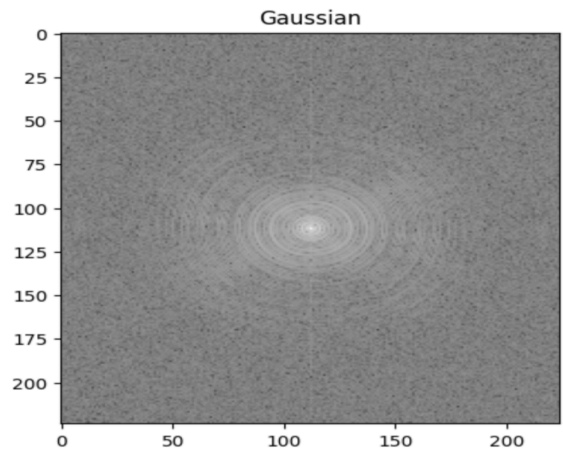

*Source: output of FFT and FFTShift functions*

*Figure 3: accuracy and Loss charts of the VGG16 model for noise type detection in the frequency domain*

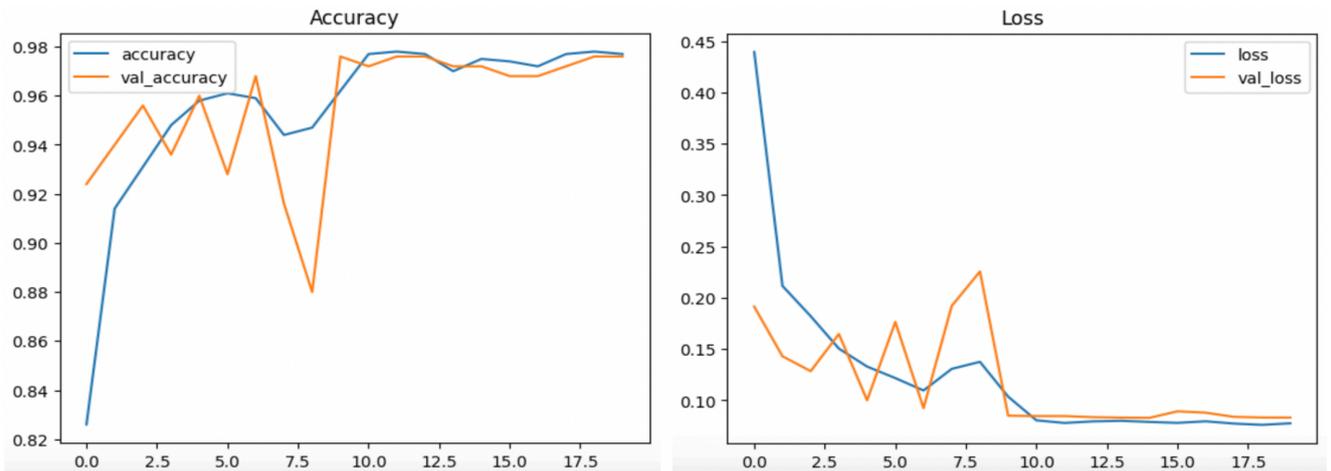

*Source: output of the VGG16 model*

### 2.3 denoising framework

periodic noise in an image is from electrical interference during the image-capturing process. An image affected by periodic noise will look like a repeating pattern has been added on top of the original image. In the frequency domain, this type of noise can be seen as discrete spikes.[11]

Also, Principal sources of Gaussian noise in digital images arise during acquisition. The sensor has inherent noise due to the level of illumination and its temperature, and the electronic circuits connected to the sensor inject their share of electronic circuit noise.[12]

Figure 4: Schematic of proposed Autoencoder architecture

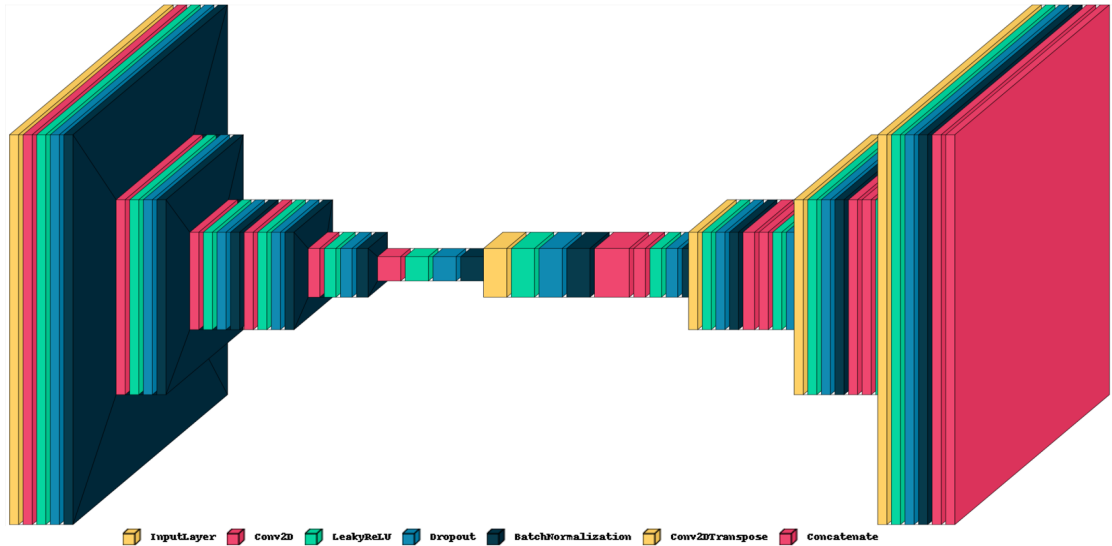

Source: this image is created by visualkeras tool, GitHub repository.
https://github.com/paulgavrikov/visualkeras

images denoising is a common application of autoencoders. Noise in images can be understood as a *random variation in the color or brightness of images*, degrading their quality. Convolutional Autoencoders can be used for this purpose. The encoder learns to extract the features separating them from the noise in the image. Thereby compressing the image. The final compressed representations from the bottleneck are passed over to the decoder. The decoder finally decompresses the image minimizing the noise.[13]

In our proposed architecture, skip connections are established between corresponding encoder and decoder layers at strategic positions. These connections enable the network to retain high-resolution information from earlier layers, preserving fine details during the up-sampling process. Additionally, skip connections facilitate the training process by mitigating the vanishing gradient problem, ensuring smoother gradient flow throughout the network.[14]

The count and position of skip connections can significantly impact the overall performance of the model. In our case study, we examined various scenarios based on our inductive bias and achieved excellent quality. [14]

### 2.3.1 Preprocessing

We split our dataset into training and test sets, allocating approximately 85% for training and 15% for testing. Subsequently, we randomly selected images to populate each set, resulting in 1032 images for training and 182 images for testing. Additionally, since the input dimensions of our chosen neural network are 256 x 256, we resized all images accordingly.

Following resizing, we applied a normalization operation to standardize the pixel values across the dataset.

### 2.3.2 model selection and training

Model selection is critical because it directly impacts the performance, generalization ability, scalability, and interpretability of the deep learning system for the given task and constraints. For example, in our case, while the Autoencoder deep model effectively removes both periodic and Gaussian noise, it fails to adequately enhance images affected by salt and pepper noise. Additionally, the median filter can only perform well on salt and pepper noise. As a solution, we implemented three sequential median filters with a kernel size of 3, based on the fact that the median filter is non-linear (order filter), and we can apply this filter repeatedly on the image. Other kernel sizes (5, 7, 9) yielded inferior results.

*Figure 5: median filter formula*

$$s(x,y) = \underset{(i,j) \in A_{xy}}{median} \{r(i,j)\}$$

**source:** Meduim.com, Remove Salt and Pepper noise with Median Filtering

### 2.3.3 model evaluation

We assess the performance of our models based on four metrics: SSIM, PSNR, LPIPS, and visual quality, the latter being a subjective metric. For this assessment, we selected a random 25% (approximately 100 images) of the second phase dataset to evaluate the denoising process. We have used AlexNet weights to calculate these metrics in Pytorch.

*Table 2: denoiser model quantitative comparison based on standard metrics*

| Model | Usage | Avg SSIM | Avg PSNR | Avg LPIPS | MAX SSIM | MAX PSNR | MAX PSNR |
|---|---|---|---|---|---|---|---|
| Sequential Median filters | Salt & pepper noise | 87.96 ±1.50 | 34.65 ±0.48 | 5.84 ±1.26 | 91.38 | 36.06 | 10.85 |
| Proposed AutoEncoder | Gaussian noise | 64.04 ±12.88 | 23.73 ±1.73 | 18.98 ±10.68 | 88.32 | 27.75 | 46.76 |
| Proposed AutoEncoder | Periodic noise | 81.69 ±4.75 | 22.83 ±3.04 | 5.12 ±2.14 | 87.54 | 26.82 | 13.03 |

*Source: outputs of models*

*All results are based on random sampling from Dataset 2 of 25% of the instances.*

*Figure 6: Gaussian noise removal using Autoencoder (PSNR=27.46)*

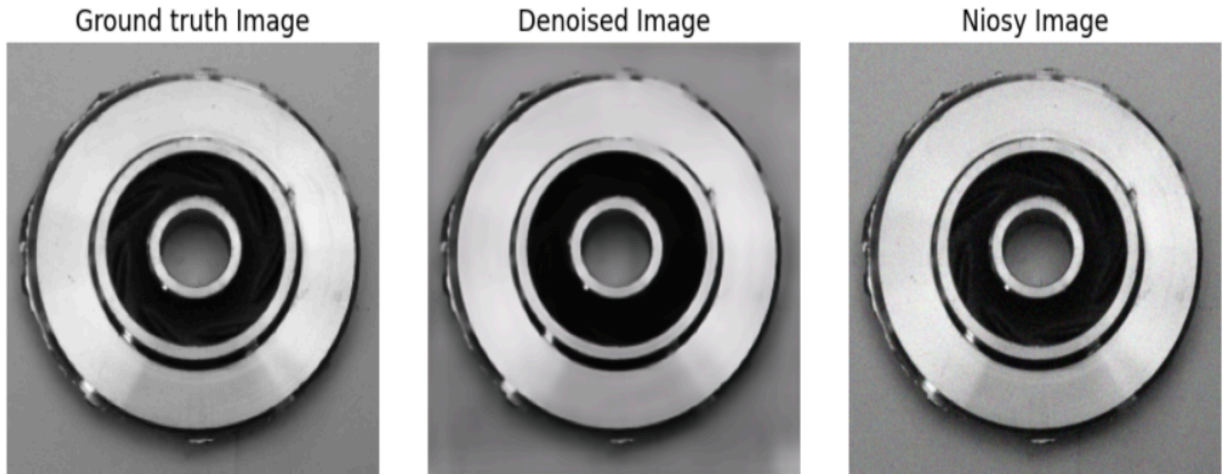

*Figure 7: Periodic noise removal using Autoencoder (PSNR=26.43)*

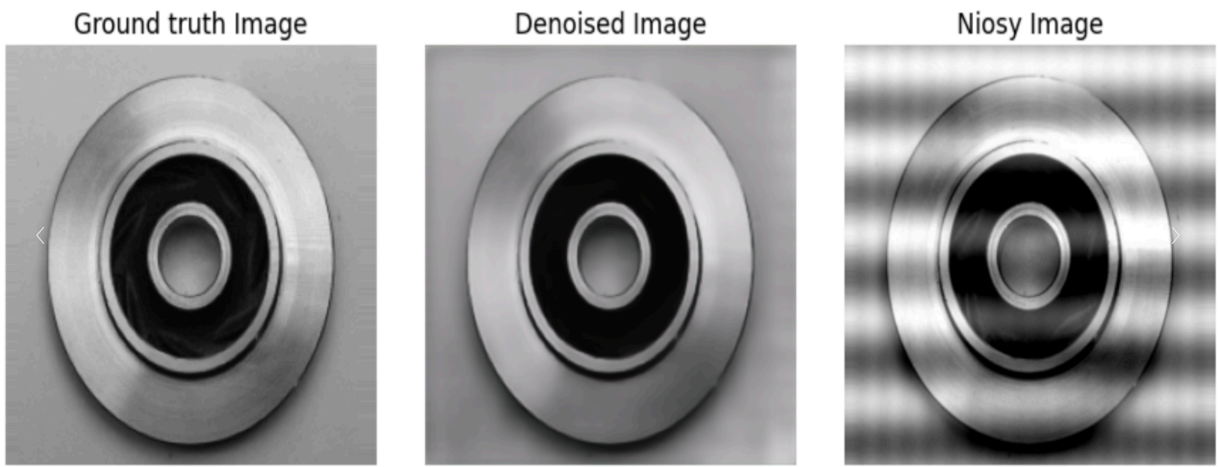

*Figure 8: Salt and pepper noise removal using median filter (PSNR=35.22)*

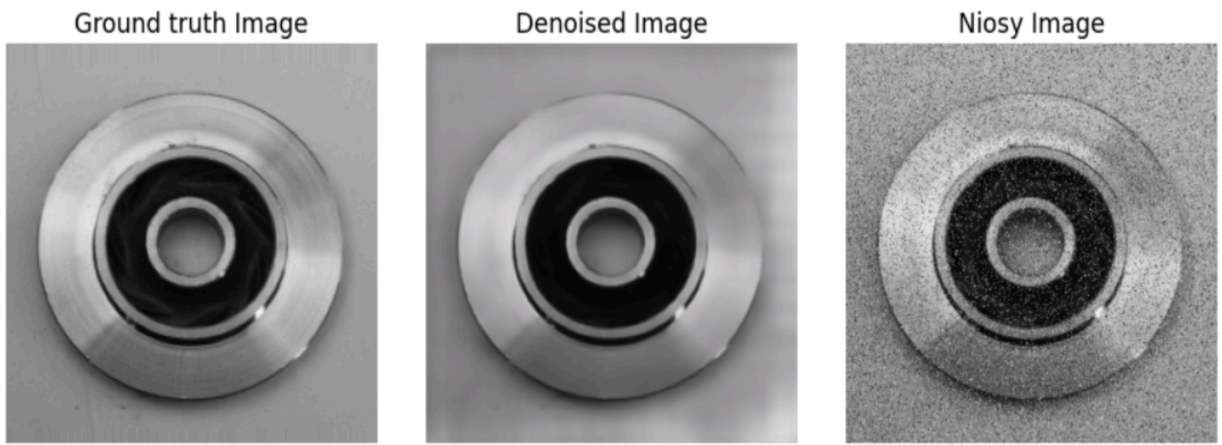

## 3 Achievements

*Figure 9: VGG16 performance improvement in binary classification comparison*

```
classification using VGG16 before noise removal
3/3 [==============================] - 1s 132ms/step - loss
Test Accuracy: 94.62365508079529
3/3 [==============================] - 1s 125ms/step

[INFO] Confusion Matrix:
[[32 24]
 [23 14]]

[INFO] classification report:
              precision    recall  f1-score   support

    Defected       0.58      0.57      0.58        56
          OK       0.37      0.38      0.37        37

    accuracy                           0.49        93
   macro avg       0.48      0.47      0.47        93
weighted avg       0.50      0.49      0.50        93

Sensitivity: 0.3783783783783784
Specificity: 0.5714285714285714

classification using VGG16 after noise removal
4/4 [==============================] - 3s 724ms/step - loss: 0.0969
Test Accuracy: 97.02970385551453
4/4 [==============================] - 1s 126ms/step

[INFO] Confusion Matrix:
[[40 23]
 [26 12]]

[INFO] classification report:
              precision    recall  f1-score   support

    Defected       0.61      0.63      0.62        63
          OK       0.34      0.32      0.33        38

    accuracy                           0.51       101
   macro avg       0.47      0.48      0.47       101
weighted avg       0.51      0.51      0.51       101

Sensitivity: 0.3157894736842105
Specificity: 0.6349206349206349
```

*Figure 10: ROC chart of VGG16's performance improvement in binary classification*

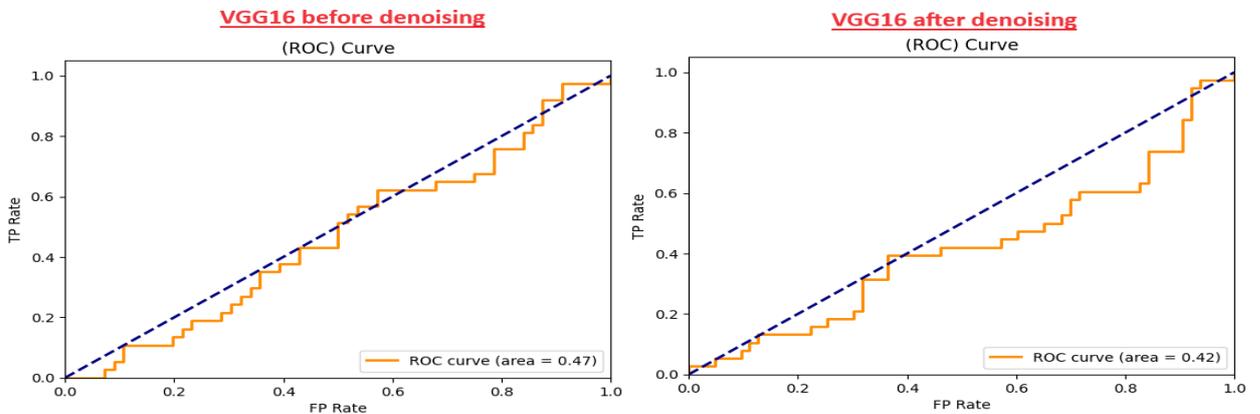

*Figure 11: InceptionV3 performance improvement in binary classification comparison*

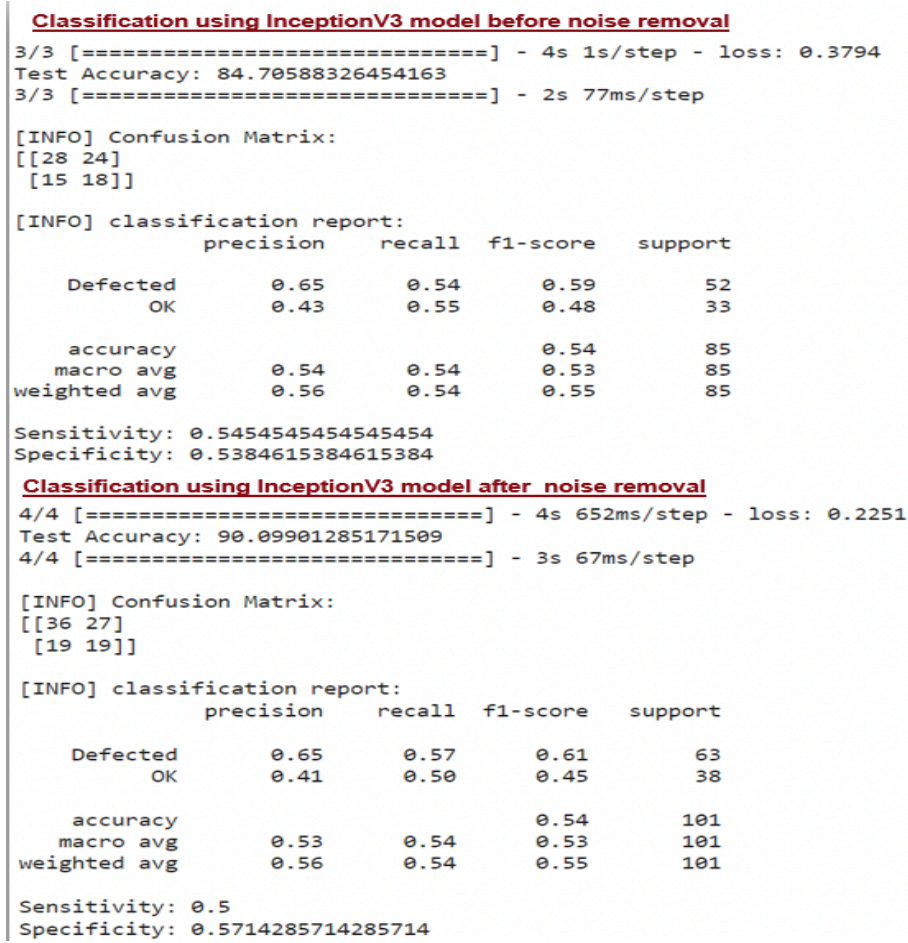

*Figure 12: ROC chart of InceptionV3's performance improvement in binary classification*

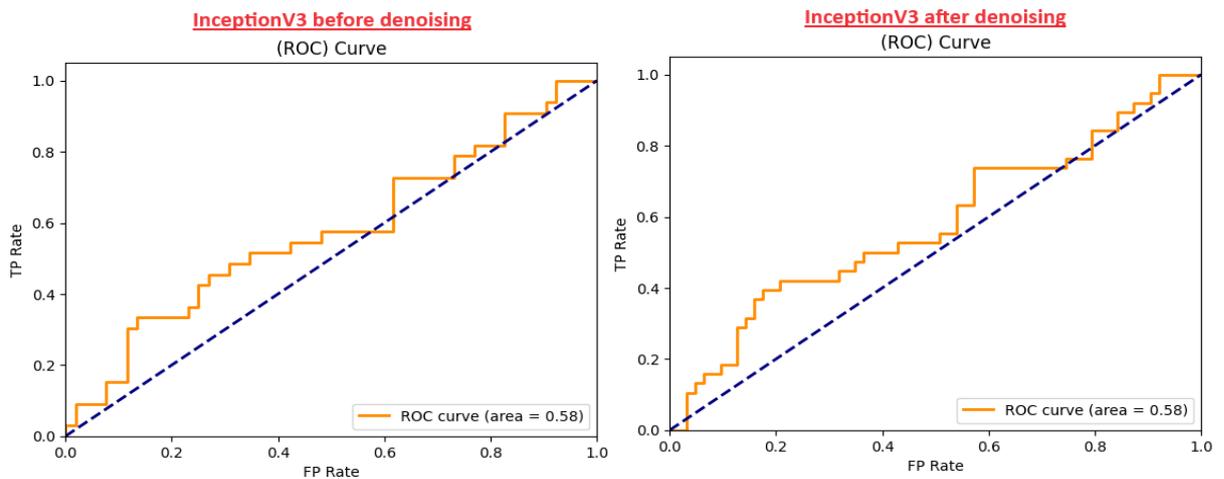

## 4 Conclusion and discussion

In our study, we meticulously trained and compared specialized deep learning models tailored to different casting product binary classification aspects. By carefully selecting and refining models like VGG16, AlexNet, CNN, InceptionV3, and ResNet50, we aimed to find the best-performing ones with high accuracy and the ability to generalize well to new data, crucial for real-world applications.

Expanding our investigation, we explored noise characterization using advanced architectures like VGG16 and Inception-ResNet-V2 in the frequency domain. This helped us understand how different types of noise affect classification performance, guiding our efforts to improve accuracy through noise reduction techniques.

Additionally, we identified an optimal point for our model. Given the complexity of our model, excessive application may lead to the removal of crucial image details, resulting in poor image quality despite favorable metrics. This could manifest as blurring if the model is applied multiple times or if the model itself is overly complex

To combat the negative effects of noise on classification, we employed sophisticated autoencoder-based denoising methods alongside median filters. This comprehensive approach, tailored to the specific noise profiles encountered, significantly improved both accuracy and generalization in defect detection.

At each stage of our analysis, we aimed for sophistication and precision. By combining empirical examination with methodological refinement, we contribute to advancing deep learning methods for industrial image analysis, paving the way for more resilient and adaptable computational systems.

**References**

[1] Simonyan, K., & Zisserman, A. (2014). Very deep convolutional networks for large-scale image recognition. arXiv preprint arXiv:1409.1556.

[2] He, K., Zhang, X., Ren, S., & Sun, J. (2016). Deep residual learning for image recognition. In Proceedings of the IEEE conference on computer vision and pattern recognition (CVPR) (pp. 770-778).

[3] Szegedy, C., Vanhoucke, V., Ioffe, S., Shlens, J., & Wojna, Z. (2016). Rethinking the Inception architecture for computer vision. In Proceedings of the IEEE conference on computer vision and pattern recognition (CVPR) (pp. 2818-2826).

[4] Krizhevsky, A., Sutskever, I., & Hinton, G. E. (2012). ImageNet Classification with Deep Convolutional Neural Networks. In *Advances in Neural Information Processing Systems*, 25.

[5] Szegedy, C., Ioffe, S., & Vanhoucke, V. (2017). Inception-v4, Inception-ResNet and the impact of residual connections on learning. In Proceedings of the AAAI conference on artificial intelligence.

[6] Chollet, F. et al. (n.d.). Keras Documentation: Callbacks Module. Retrieved from https://keras.io/api/callbacks/

[7] Wang, L., & Sun, Y. (2022). Image classification using convolutional neural network with wavelet domain inputs. IET Image Processing, 16, 2037–2048. https://doi.org/10.1049/ipr2.12466

[8] Lehar, S. (n.d.). An intuitive explanation of Fourier theory. Retrieved from http://cns-alumni.bu.edu/~slehar/fourier/fourier.html

[9] Medium.com. (2020, June 11). Study of state-of-the-art image classification models and their application to face recognition. Analytics Vidhya. Retrieved from https://medium.com/analytics-vidhya/study-of-state-of-the-art-image-classification-models-and-their-application-to-face-recognition-bdd6b3820ac

[10] Mascarenhas, S., & Agarwal, M. (2021). A comparison between VGG16, VGG19 and ResNet50 architecture frameworks for Image Classification. In 2021 International Conference on Disruptive Technologies for Multi-Disciplinary Research and Applications (CENTCON) (pp. 96-99). doi:10.1109/CENTCON52345.2021.9687944

[11] Gonzalez, R. C., & Woods, R. E. (2007). Digital Image Processing. Pearson Prentice Hall. ISBN 978-0-13-168728-8.

[12] Cattin, P. (2012, April 24). Image Restoration: Introduction to Signal and Image Processing. MIAC, University of Basel. Retrieved from https://web.archive.org/web/20160918213602/https://www.miac.unibas.ch/fileadmin/user_upload/miac/dam/MIAC_Images/PDFs/Vorlesungen/ImageRestoration_2012.pdf

[13] Bajaj, K., Singh, D. K., & Ansari, M. A. (2020). Autoencoders Based Deep Learner for Image Denoising. *Procedia Computer Science, 171*, 1535-1541. Third International Conference on Computing and Network Communications (CoCoNet'19). https://doi.org/10.1016/j.procs.2020.04.164

*[14] Benmeziane, H., Ounnoughene, A. Z., Hamzaoui, I., & Bouhadjar, Y. (2023). Skip Connections in Spiking Neural Networks: An Analysis of Their Effect on Network Training. arXiv preprint arXiv:2303.13563. https://doi.org/10.48550/arXiv.2303.13563*